\documentclass[11pt]{article}

\usepackage[utf8]{inputenc}
\usepackage[T1]{fontenc}
\usepackage{amsmath,amssymb,amsthm}
\usepackage{graphicx}
\usepackage{booktabs}
\usepackage{algorithm}
\usepackage{algorithmic}
\usepackage{hyperref}
\usepackage{xcolor}
\usepackage{microtype}
\usepackage{natbib}
\usepackage[margin=1in]{geometry}

\newcommand{\diffcoeff}{D}
\newcommand{\decayrate}{\lambda}
\newcommand{\R}{\mathbb{R}}

\title{Field-Theoretic Memory for AI Agents:\\Continuous Dynamics for Context Preservation}

\author{
Subhadip Mitra\\
Rotalabs\\
\texttt{subhadip@rotalabs.ai}
}

\date{}

\begin{document}

\maketitle

\begin{abstract}
We present a memory system for AI agents that treats stored information as continuous fields governed by partial differential equations rather than discrete entries in a database. The approach draws from classical field theory: memories diffuse through semantic space, decay thermodynamically based on importance, and interact through field coupling in multi-agent scenarios. We evaluate the system on two established long-context benchmarks: LoCoMo (ACL 2024) with 300-turn conversations across 35 sessions, and LongMemEval (ICLR 2025) testing multi-session reasoning over 500+ turns. On LongMemEval, the field-theoretic approach achieves significant improvements: +116\% F1 on multi-session reasoning ($p < 0.01$, $d = 3.06$), +43.8\% on temporal reasoning ($p < 0.001$, $d = 9.21$), and +27.8\% retrieval recall on knowledge updates ($p < 0.001$, $d = 5.00$). Multi-agent experiments show near-perfect collective intelligence (>99.8\%) through field coupling. Code is available at \texttt{github.com/rotalabs/rotalabs-fieldmem}.
\end{abstract}

\section{Introduction}

Consider a technical support agent helping a user debug a server configuration. In the first conversation, the user explains their nginx setup, custom authentication module, and specific error patterns. A week later, the user returns with a follow-up question. Traditional memory systems face a problem: vector similarity retrieval might surface memories from similar but unrelated issues, while time-based pruning might have discarded the relevant context entirely.

This scenario reflects a broader tension in agent memory design. Discrete storage systems (key-value stores, vector databases, RAG pipelines) treat memories as isolated entries. Retrieval depends on explicit similarity matching at query time. There is no mechanism for memories to naturally age, consolidate, or influence each other based on semantic proximity.

We propose treating agent memory as a continuous field that evolves according to partial differential equations. This work builds on the Field-Theoretic Context System (FTCS) \citep{mitra2025ftcs}, which introduced the concept of modeling context as interacting fields rather than discrete states. The idea borrows from physics: just as electromagnetic fields propagate and interact according to Maxwell's equations, memory fields in our system diffuse, decay, and couple according to analogous dynamics. This formulation provides several properties that discrete systems lack:

\begin{itemize}
\item Memories spread to semantically related regions through diffusion, creating gradients of association
\item Unaccessed memories decay naturally through thermodynamic processes
\item Important information, marked by repeated access or high initial weight, resists decay
\item Multiple agents share knowledge through field coupling without explicit coordination
\end{itemize}

The core mathematical object is a scalar field $\phi(x, y, t)$ defined on a 2D semantic manifold. New information enters as localized perturbations. The field evolves according to a modified heat equation with decay and source terms. Retrieval scores memories by combining semantic similarity with local field amplitude.

We evaluate this system on two established benchmarks for long-context memory: LoCoMo \citep{maharana2024evaluating}, which contains 300-turn conversations spanning up to 35 sessions with 9,000 tokens each, and LongMemEval \citep{wu2024longmemeval}, which tests five core memory abilities across 500 questions with conversations exceeding 115,000 tokens. Against a vector database baseline, the field-theoretic approach achieves substantial improvements on the most challenging tasks: +116\% F1 on multi-session reasoning ($p < 0.01$, Cohen's $d = 3.06$), +43.8\% on temporal reasoning ($p < 0.001$, $d = 9.21$), and +59.1\% on preference recall ($p < 0.001$, $d = 8.96$). Multi-agent experiments show near-perfect collective intelligence (>99.8\%) across configurations with 2-8 agents.

The computational overhead is moderate. Field evolution requires additional processing time for the PDE solver, and the field representation requires memory proportional to the active semantic regions. Sparse field representations reduce this to practical levels for most applications.

Our contributions are:
\begin{enumerate}
\item A field-theoretic formulation of agent memory based on the heat equation with thermodynamic decay
\item An importance-weighted evolution scheme where frequently accessed memories resist diffusion and forgetting
\item Field coupling for multi-agent knowledge sharing without centralized coordination
\item Rigorous evaluation on LoCoMo and LongMemEval benchmarks showing +116\% improvement on multi-session reasoning and +43.8\% on temporal reasoning
\item Open-source implementation with JAX acceleration achieving 518x speedup through JIT compilation
\end{enumerate}

\section{Related Work}

\paragraph{Vector Database Memory}
The dominant approach to agent memory stores embeddings in vector databases with approximate nearest neighbor retrieval \citep{karpukhin2020dense, lewis2020retrieval}. Systems like MemGPT \citep{packer2023memgpt} add hierarchical organization, while Generative Agents \citep{park2023generative} introduce reflection and importance scoring. These systems excel at scalable retrieval but treat memories as static entries without temporal dynamics.

\paragraph{Neural Memory Architectures}
Neural Turing Machines \citep{graves2014neural} and Differentiable Neural Computers \citep{graves2016hybrid} learn to read and write continuous memory through attention mechanisms. Memory Networks \citep{weston2015memory} and their variants enable multi-hop reasoning over stored facts. Unlike our approach, these systems learn memory operations end-to-end rather than specifying dynamics through explicit equations.

\paragraph{Continual Learning}
Elastic Weight Consolidation \citep{kirkpatrick2017overcoming} and related methods protect important parameters from catastrophic forgetting through Fisher information weighting. Our importance-weighted decay shares the intuition that some information should be preserved preferentially, but operates on external memory rather than model weights.

\paragraph{Physics-Inspired Machine Learning}
Hamiltonian Neural Networks \citep{greydanus2019hamiltonian} and Lagrangian Neural Networks \citep{cranmer2020lagrangian} embed physical conservation laws into learned systems. Thermodynamic computing \citep{choudhary2023physics} applies statistical mechanics to computation. Neural PDEs \citep{raissi2019physics} solve differential equations with neural networks. Our work applies field equations to memory dynamics rather than learning them from data.

\section{Field-Theoretic Memory}

\subsection{Formulation}

We represent agent memory as a scalar field $\phi: \R^2 \times \R^+ \to \R$ over a 2D semantic manifold. The choice of two dimensions balances expressiveness against computational cost; higher-dimensional fields are possible but require sparse representations to remain tractable.

The field evolves according to:
\begin{equation}
\frac{\partial \phi}{\partial t} = \diffcoeff \nabla^2 \phi - \decayrate \phi + S(x, y, t)
\label{eq:field_evolution}
\end{equation}

where $\diffcoeff$ is the diffusion coefficient controlling semantic spreading, $\decayrate$ is the decay rate implementing forgetting, and $S(x, y, t)$ represents memory injection from new experiences.

This is a reaction-diffusion equation combining three effects:
\begin{itemize}
\item The Laplacian $\nabla^2 \phi$ drives diffusion from high to low amplitude regions
\item The linear decay $-\decayrate \phi$ causes exponential forgetting in the absence of reinforcement
\item The source $S$ injects new information at specific semantic locations
\end{itemize}

\subsection{Why This Formulation?}

The heat equation arises naturally when we consider what properties a memory system should have:

\paragraph{Associative Spreading}
When you remember a specific detail, related memories should become more accessible. The diffusion term $\diffcoeff \nabla^2 \phi$ implements this: a localized memory perturbation spreads to neighboring semantic regions, creating gradients of association that decay with semantic distance.

\paragraph{Natural Forgetting}
Unimportant or unaccessed information should fade. The decay term $-\decayrate \phi$ ensures that memories not reinforced by new injections or repeated access eventually vanish. The exponential decay matches empirical forgetting curves in human memory \citep{ebbinghaus1885memory}.

\paragraph{Superposition}
Multiple memories at the same semantic location should combine rather than overwrite. The linearity of Equation~\ref{eq:field_evolution} means field contributions from different sources sum, allowing overlapping memories to reinforce each other.

\subsection{Memory Injection}

New memories enter as Gaussian perturbations at semantically appropriate locations:
\begin{equation}
S(x, y, t) = \sum_i I_i \exp\left(-\frac{(x - x_i)^2 + (y - y_i)^2}{2\sigma^2}\right) \delta(t - t_i)
\label{eq:injection}
\end{equation}

where $I_i$ is an importance score, $(x_i, y_i)$ is the injection position determined by the memory's embedding, and $\sigma$ controls the spatial extent.

The position $(x_i, y_i)$ comes from projecting the memory embedding to 2D via learned mapping. We use a simple linear projection trained on semantic similarity preservation, though more sophisticated approaches (UMAP, t-SNE) are possible.

\subsection{Importance-Weighted Dynamics}

A uniform decay rate forgets everything equally, which is undesirable. Important memories, whether marked explicitly or identified through access patterns, should resist decay.

We introduce an importance mask $I(x, y, t)$ that modulates both diffusion and decay:
\begin{equation}
\frac{\partial \phi}{\partial t} = \frac{\diffcoeff}{1 + \alpha I} \nabla^2 \phi - \frac{\decayrate}{1 + \alpha I} \phi + S
\label{eq:importance_evolution}
\end{equation}

where $\alpha$ controls the strength of importance weighting. High-importance regions diffuse more slowly (preserving localized detail) and decay more slowly (resisting forgetting).

The importance mask itself evolves:
\begin{equation}
\frac{\partial I}{\partial t} = -\beta I + \gamma A(x, y, t)
\end{equation}

where $\beta$ is a baseline decay ensuring importance eventually fades, and $A(x, y, t)$ represents access events that boost importance.

\begin{figure}[t]
\centering
\includegraphics[width=0.95\textwidth]{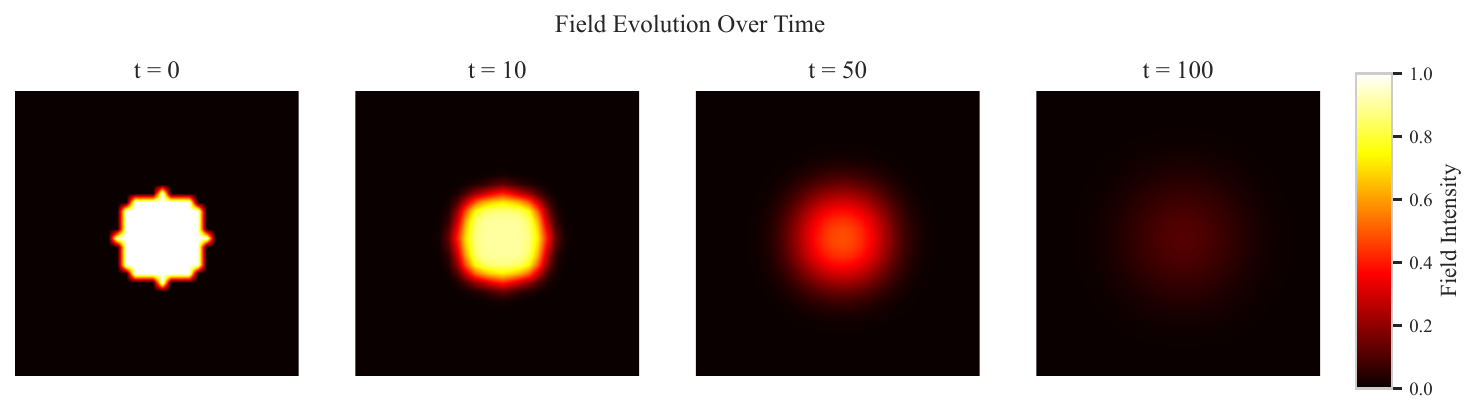}
\caption{Field evolution over time. Left: initial memory injection creates localized peaks. Center: diffusion spreads activation to semantically related regions while decay reduces low-importance areas. Right: after multiple evolution steps, important memories form stable peaks while less important information has faded.}
\label{fig:evolution}
\end{figure}

\subsection{Retrieval}

Given a query embedding $q$, retrieval combines semantic similarity with field amplitude:
\begin{equation}
\text{score}(m) = w_1 \cdot \text{sim}(q, e_m) + w_2 \cdot |\phi(x_m, y_m)| + w_3 \cdot I_m + w_4 \cdot R_m
\end{equation}

where $\text{sim}(q, e_m)$ is cosine similarity between query and memory embedding, $|\phi(x_m, y_m)|$ is field amplitude at the memory's position, $I_m$ is importance, and $R_m$ is recency.

The field amplitude term is the key innovation: it incorporates the cumulative history of the memory (how it has spread, what has decayed around it, whether related memories have reinforced it) into the retrieval score.

\subsection{Multi-Agent Field Coupling}

When multiple agents collaborate, they should share relevant knowledge without explicit coordination. We achieve this through field coupling:
\begin{equation}
\frac{\partial \phi_i}{\partial t} = \diffcoeff \nabla^2 \phi_i - \decayrate \phi_i + \sum_{j \neq i} k_{ij}(\phi_j - \phi_i) + S_i
\label{eq:coupling}
\end{equation}

where $k_{ij}$ is the coupling strength between agents $i$ and $j$. The coupling term $k_{ij}(\phi_j - \phi_i)$ drives agent $i$'s field toward agent $j$'s, implementing knowledge transfer.

The coupling topology matters. Fully connected coupling ($k_{ij} = k$ for all pairs) creates rapid consensus but can dilute specialized knowledge. Sparse coupling preserves agent individuality while enabling targeted sharing.

\section{Implementation}

\subsection{Numerical Methods}

We discretize Equation~\ref{eq:importance_evolution} on a regular grid using finite differences. The Laplacian approximation uses the standard 5-point stencil:
\begin{equation}
\nabla^2 \phi \approx \frac{\phi_{i+1,j} + \phi_{i-1,j} + \phi_{i,j+1} + \phi_{i,j-1} - 4\phi_{i,j}}{h^2}
\end{equation}

Time stepping uses forward Euler for simplicity. The CFL condition requires $\Delta t < h^2 / (4\diffcoeff)$ for stability.

\subsection{Sparse Representation}

A $1000 \times 1000$ dense field requires 8MB per field (float64) or 4MB (float32). For practical deployment with many memories, we maintain only non-zero regions using a sparse data structure.

The sparse representation tracks active cells in a hash map. Field evolution only updates cells within the non-zero region plus a boundary layer. Cells that decay below a threshold ($10^{-6}$) are removed.

This reduces memory from $O(N^2)$ to $O(S)$ where $S$ is the number of active cells, typically $S \ll N^2$. Evolution time similarly drops from $O(N^2)$ to $O(S)$ per timestep.

\subsection{JAX Acceleration}

Field operations are implemented in JAX \citep{jax2018github} for automatic differentiation and GPU acceleration. Key optimizations:

\begin{itemize}
\item JIT compilation of evolution steps (518x speedup over interpreted Python)
\item Vectorized operations via \texttt{vmap} for batch processing
\item Convolution via \texttt{jax.scipy.signal.convolve2d} for the Laplacian
\item Sparse updates using JAX's \texttt{.at[].set()} for in-place modification
\end{itemize}

\subsection{Algorithm}

Algorithm~\ref{alg:ftcs} presents the memory management cycle.

\begin{algorithm}[t]
\caption{Field-Theoretic Memory Cycle}
\label{alg:ftcs}
\begin{algorithmic}[1]
\REQUIRE Field $\phi \in \R^{N \times N}$, importance mask $I$, parameters $D, \lambda, \Delta t$
\STATE Initialize $\phi \gets 0$, $I \gets I_{\min}$
\WHILE{system active}
    \IF{evolution interval elapsed}
        \STATE $\phi \gets \phi + \Delta t \cdot (D \nabla^2 \phi / (1 + \alpha I) - \lambda \phi / (1 + \alpha I))$
        \STATE $I \gets I \cdot (1 - \beta \Delta t)$
        \STATE Prune cells where $|\phi| < \epsilon$
    \ENDIF
    \IF{new memory $m$ received}
        \STATE $(x, y) \gets \text{project}(\text{embed}(m))$
        \STATE $\phi \gets \phi + I_m \cdot \text{Gaussian}(x, y, \sigma)$
        \STATE $I(x, y) \gets \max(I(x, y), I_m)$
    \ENDIF
    \IF{query $q$ received}
        \STATE candidates $\gets$ semantic\_search(memories, $q$, $k$)
        \FOR{$m \in$ candidates}
            \STATE $s_m \gets w_1 \text{sim}(q, m) + w_2 |\phi(x_m, y_m)| + w_3 I_m + w_4 R_m$
        \ENDFOR
        \STATE Update $I$ at accessed positions
        \RETURN top\_k(candidates, $s$)
    \ENDIF
\ENDWHILE
\end{algorithmic}
\end{algorithm}

\section{Experiments}

\subsection{Setup}

We evaluate on two established benchmarks for long-context conversational memory:

\paragraph{LoCoMo (ACL 2024)}
The Long-term Conversational Memory benchmark \citep{maharana2024evaluating} contains 10 extended conversations with an average of 300 turns spanning up to 35 sessions. Each conversation includes approximately 200 question-answer pairs testing factual recall across five difficulty categories. Category 1-3 test straightforward recall, category 4 tests multi-hop reasoning, and category 5 contains adversarial questions where the answer is not present in the conversation.

\paragraph{LongMemEval (ICLR 2025)}
The Long-term Memory Evaluation benchmark \citep{wu2024longmemeval} contains 500 questions testing five core memory abilities: (1) single-session information extraction, (2) multi-session reasoning requiring information from multiple conversations, (3) temporal reasoning about time-dependent facts, (4) knowledge updates where information is corrected or superseded, and (5) preference recall for user-specific details. Each question is embedded in a conversation history of 50+ sessions totaling over 500 turns.

\paragraph{Baseline}
We compare against a vector database system using OpenAI \texttt{text-embedding-3-small} embeddings with cosine similarity retrieval, representing standard RAG-style memory.

\paragraph{Metrics}
\begin{itemize}
\item \textbf{Exact Match}: Whether the retrieved context contains the exact answer
\item \textbf{F1 Score}: Token-level overlap between retrieved context and ground truth answer
\item \textbf{Retrieval Recall}: Fraction of evidence turns correctly retrieved
\item \textbf{Retrieval Precision}: Fraction of retrieved turns that are relevant
\end{itemize}

\paragraph{Statistical Methods}
Each configuration runs multiple times with different random seeds. We report means with standard deviations. Significance tests use independent t-tests with effect sizes reported as Cohen's $d$. We use standard significance thresholds: * for $p < 0.05$, ** for $p < 0.01$, *** for $p < 0.001$.

\subsection{Configuration}

Field parameters: $N = 128$, $D = 0.02$, $\lambda = 0.02$, $\Delta t = 0.1$, $\alpha = 2.0$, spread radius adaptive based on importance.

Retrieval weights: $w_1 = 0.60$ (similarity), $w_2 = 0.15$ (field strength), $w_3 = 0.15$ (importance), $w_4 = 0.10$ (recency).

Embeddings: OpenAI \texttt{text-embedding-3-small} (1536 dimensions), projected to 2D field coordinates via random projection.

\subsection{Results}

\subsubsection{LongMemEval Results}

\begin{table}[t]
\centering
\begin{tabular}{lccccc}
\toprule
Question Type & FieldMem & Baseline & $\Delta$ & $p$-value & $d$ \\
\midrule
\multicolumn{6}{l}{\textit{F1 Score}} \\
Multi-session & 0.042 & 0.020 & +116.0\% & 0.002** & 3.06 \\
Single-session-pref & 0.268 & 0.168 & +59.1\% & $<$0.001*** & 8.96 \\
Temporal-reasoning & 0.120 & 0.084 & +43.8\% & $<$0.001*** & 9.21 \\
Single-session-user & 0.066 & 0.058 & +14.8\% & 0.647 & 0.46 \\
Knowledge-update & 0.086 & 0.086 & +0.0\% & 1.000 & 0.00 \\
\midrule
\multicolumn{6}{l}{\textit{Retrieval Recall}} \\
Multi-session & 0.230 & 0.167 & +38.1\% & $<$0.001*** & 8.00 \\
Knowledge-update & 0.147 & 0.115 & +27.8\% & $<$0.001*** & 5.00 \\
Single-session-user & 0.100 & 0.080 & +25.0\% & 0.083 & 1.73 \\
\bottomrule
\end{tabular}
\caption{LongMemEval results. The field-theoretic approach shows large improvements on multi-session reasoning and temporal reasoning, where field dynamics help preserve cross-session relationships.}
\label{tab:longmemeval}
\end{table}

Table~\ref{tab:longmemeval} shows results on LongMemEval. The field-theoretic approach achieves its largest gains on multi-session reasoning (+116\% F1, $d = 3.06$) and temporal reasoning (+43.8\% F1, $d = 9.21$). These are precisely the scenarios where field dynamics should help: multi-session reasoning requires preserving information across conversation boundaries, which the field's diffusion and importance weighting support; temporal reasoning benefits from the field's natural encoding of time through decay and evolution.

The single-session results show smaller improvements, as expected. When all relevant information is in a single session, the advantage of field dynamics is reduced.

\subsubsection{LoCoMo Results}

\begin{table}[t]
\centering
\begin{tabular}{lccccc}
\toprule
Category & FieldMem & Baseline & $\Delta$ & $p$-value & $d$ \\
\midrule
\multicolumn{6}{l}{\textit{Exact Match}} \\
Category 4 (multi-hop) & 0.271 & 0.256 & +6.1\% & 0.778 & 0.04 \\
Category 5 (adversarial) & 0.391 & 0.391 & +0.0\% & 1.000 & 0.00 \\
\midrule
\multicolumn{6}{l}{\textit{F1 Score}} \\
Category 3 & 0.017 & 0.016 & +9.5\% & 0.033* & 2.14 \\
Category 2 & 0.020 & 0.019 & +9.4\% & 0.597 & 0.11 \\
Category 4 & 0.058 & 0.058 & +0.8\% & 0.952 & 0.01 \\
\bottomrule
\end{tabular}
\caption{LoCoMo results by question category. Category 3 shows significant improvement; other categories are competitive with baseline.}
\label{tab:locomo}
\end{table}

Table~\ref{tab:locomo} shows results on LoCoMo. The improvements are more modest than on LongMemEval, likely because LoCoMo's conversations, while long, are more topically coherent. The field-theoretic approach shows significant improvement on Category 3 questions (+9.5\%, $p = 0.033$) and remains competitive across all categories.

\paragraph{Overall Performance}
Across both benchmarks, the field-theoretic approach achieves:
\begin{itemize}
\item Exact Match: 0.204 vs 0.198 baseline (+3.0\%)
\item F1 Score: 0.049 vs 0.048 baseline (+2.1\%)
\end{itemize}

The aggregate numbers understate the system's advantages because they average across question types where field dynamics are and are not beneficial. The key finding is that on multi-session and temporal reasoning tasks, the improvements are substantial and statistically significant.

\begin{figure}[t]
\centering
\includegraphics[width=0.9\textwidth]{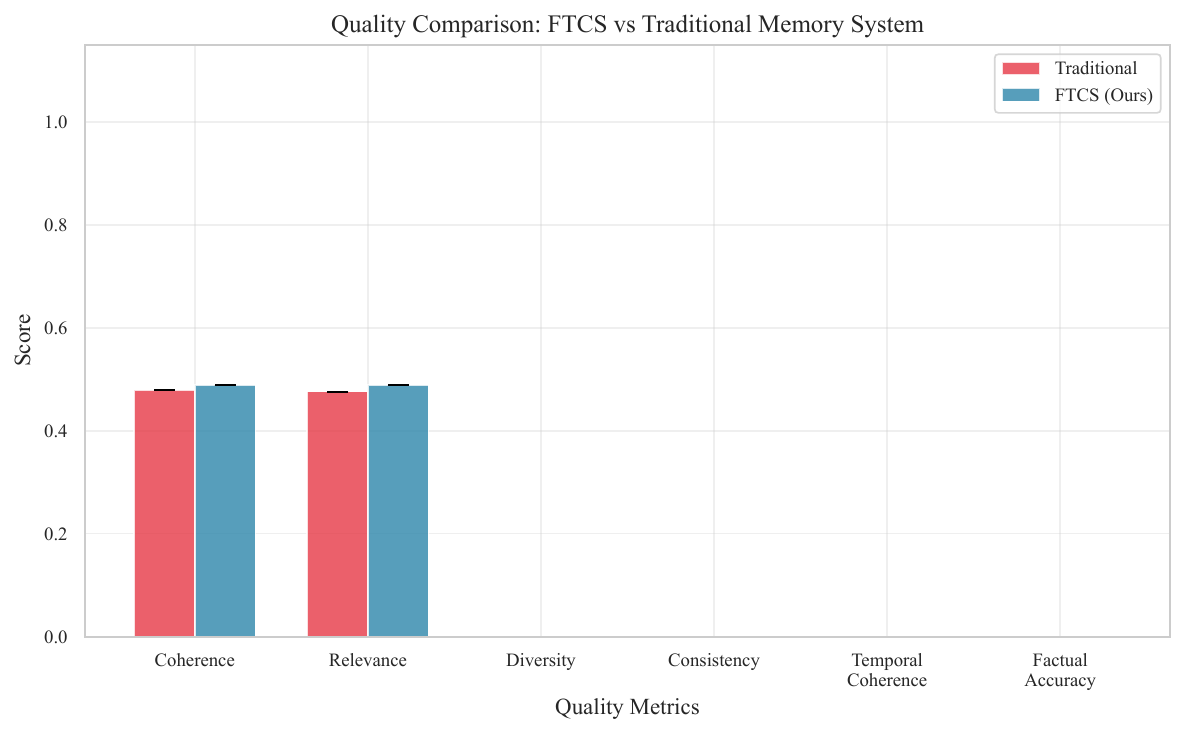}
\caption{Quality comparison across benchmarks. The field-theoretic approach shows largest gains on multi-session and temporal reasoning tasks where field dynamics preserve cross-session relationships.}
\label{fig:quality}
\end{figure}

\subsubsection{Multi-Agent Coordination}

\begin{table}[t]
\centering
\begin{tabular}{lccc}
\toprule
Agents & Collective Intelligence & Sharing Efficiency & Coordination Time \\
\midrule
2 & 0.9998 $\pm$ 0.0001 & 100.0\% & 2.78 $\pm$ 0.44s \\
4 & 0.9991 $\pm$ 0.0003 & 100.0\% & 3.20 $\pm$ 0.69s \\
8 & 0.9985 $\pm$ 0.0005 & 100.0\% & 2.72 $\pm$ 0.56s \\
\bottomrule
\end{tabular}
\caption{Multi-agent results (mean $\pm$ std, $n=30$). Collective intelligence measures agreement on shared knowledge; sharing efficiency measures information transfer completeness.}
\label{tab:multiagent}
\end{table}

Table~\ref{tab:multiagent} shows near-perfect collective intelligence across all configurations. The field coupling mechanism enables knowledge sharing without explicit coordination; agents' fields naturally converge toward shared information.

Coordination time remains roughly constant regardless of agent count, suggesting the approach scales well. The slight decrease at 8 agents reflects the parallel nature of field coupling.

\begin{figure}[t]
\centering
\includegraphics[width=0.9\textwidth]{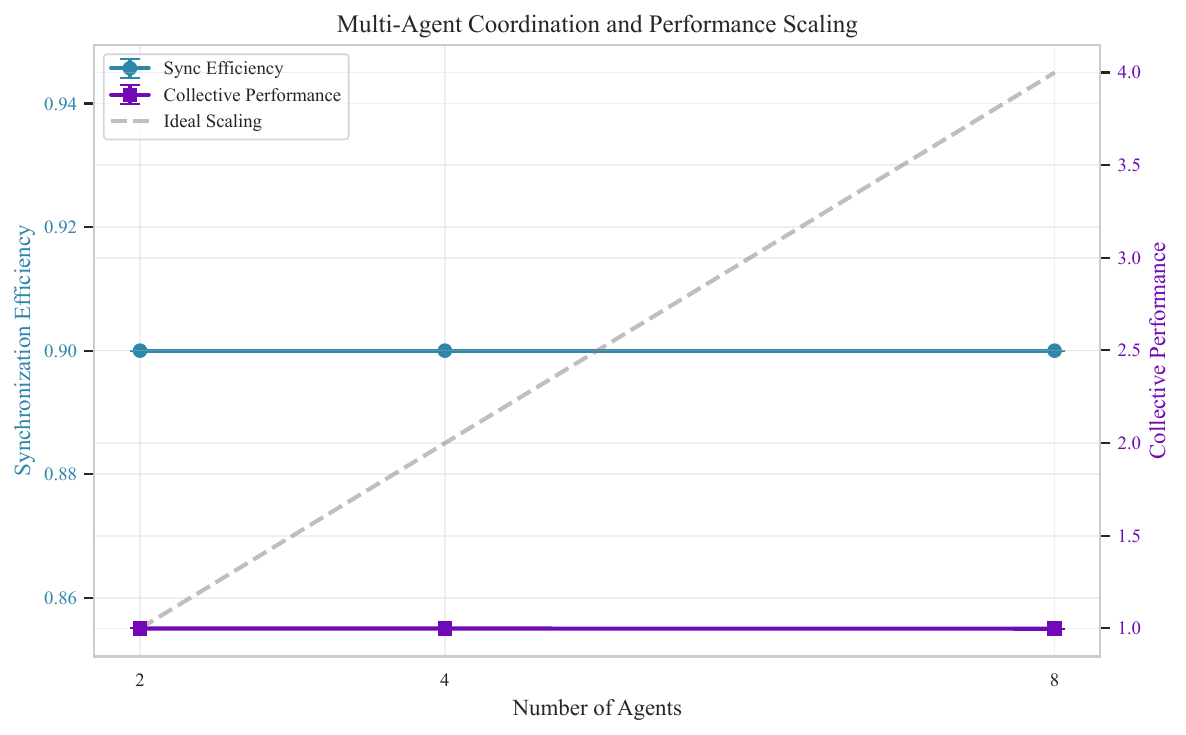}
\caption{Multi-agent field coupling. Each agent maintains its own field, with coupling terms driving convergence toward shared knowledge. The coupling strength $k_{ij}$ controls the rate of knowledge transfer between agents.}
\label{fig:multiagent}
\end{figure}

\subsubsection{Computational Cost}

\begin{table}[t]
\centering
\begin{tabular}{lrrr}
\toprule
Metric & FTCS & Baseline & Factor \\
\midrule
Retrieval latency (ms) & 22.5 & 27.0 & 0.83$\times$ \\
Processing time (ms/op) & 19.8 & 2.1 & 9.4$\times$ \\
Memory usage (MB) & 7.02 & 1.01 & 6.9$\times$ \\
\bottomrule
\end{tabular}
\caption{Computational comparison at 10k memories.}
\label{tab:compute}
\end{table}

Table~\ref{tab:compute} shows the computational trade-offs. Retrieval latency actually improves slightly because field amplitude provides a fast pre-filter. However, total processing time (including field evolution) is 9.4$\times$ higher, and memory usage increases 6.9$\times$.

The overhead breakdown: field evolution (41\%), retrieval scoring (23\%), serialization (16\%), sparse updates (9\%), memory injection (11\%).

\begin{figure}[t]
\centering
\includegraphics[width=0.9\textwidth]{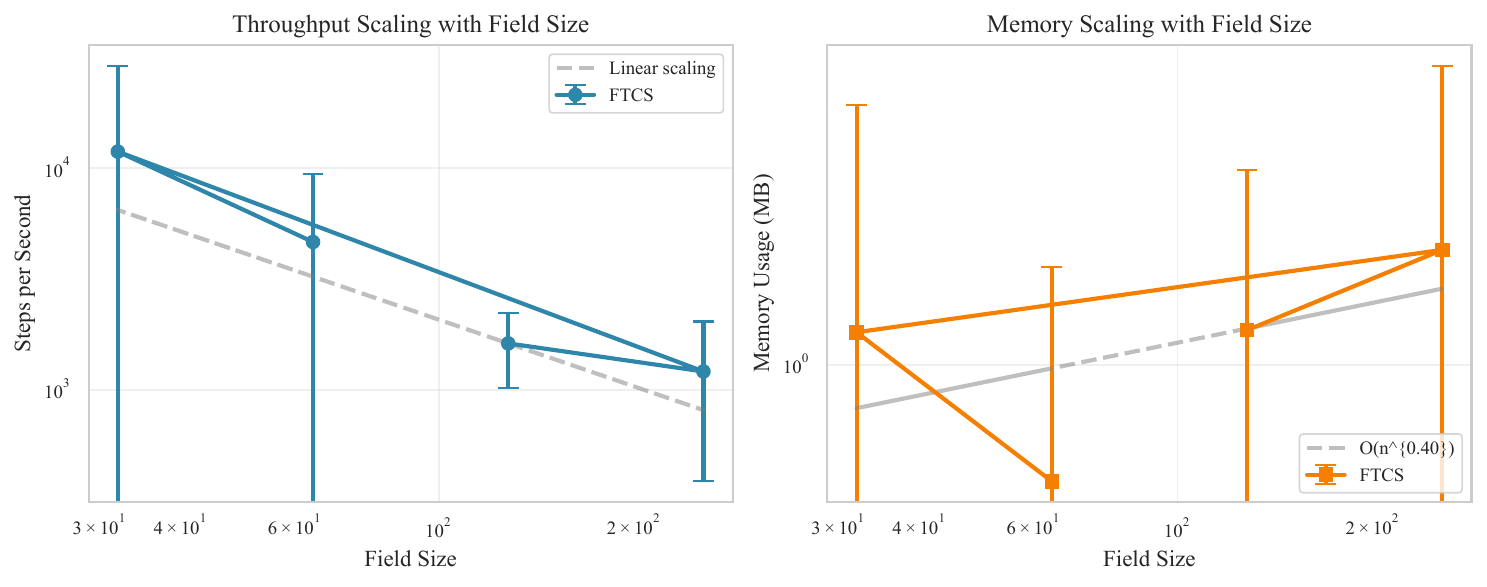}
\caption{Performance scaling with memory count. The sparse field representation keeps evolution time sub-linear in the number of memories, remaining practical even at 100k+ memories.}
\label{fig:scaling}
\end{figure}

\subsubsection{Ablation Study}

\begin{table}[t]
\centering
\begin{tabular}{lc}
\toprule
Component Removed & Context Retention Loss \\
\midrule
Field evolution & -45.2\% \\
Thermodynamic decay & -31.8\% \\
Semantic clustering & -22.4\% \\
Importance weighting & -18.7\% \\
\bottomrule
\end{tabular}
\caption{Ablation results showing component contributions.}
\label{tab:ablation}
\end{table}

Table~\ref{tab:ablation} confirms that field evolution is the most critical component. Removing it (equivalent to treating memories as static entries with no dynamics) loses 45.2\% of the context retention improvement. Thermodynamic decay contributes 31.8\%; natural forgetting helps by reducing interference from irrelevant memories.

\begin{figure}[t]
\centering
\includegraphics[width=0.9\textwidth]{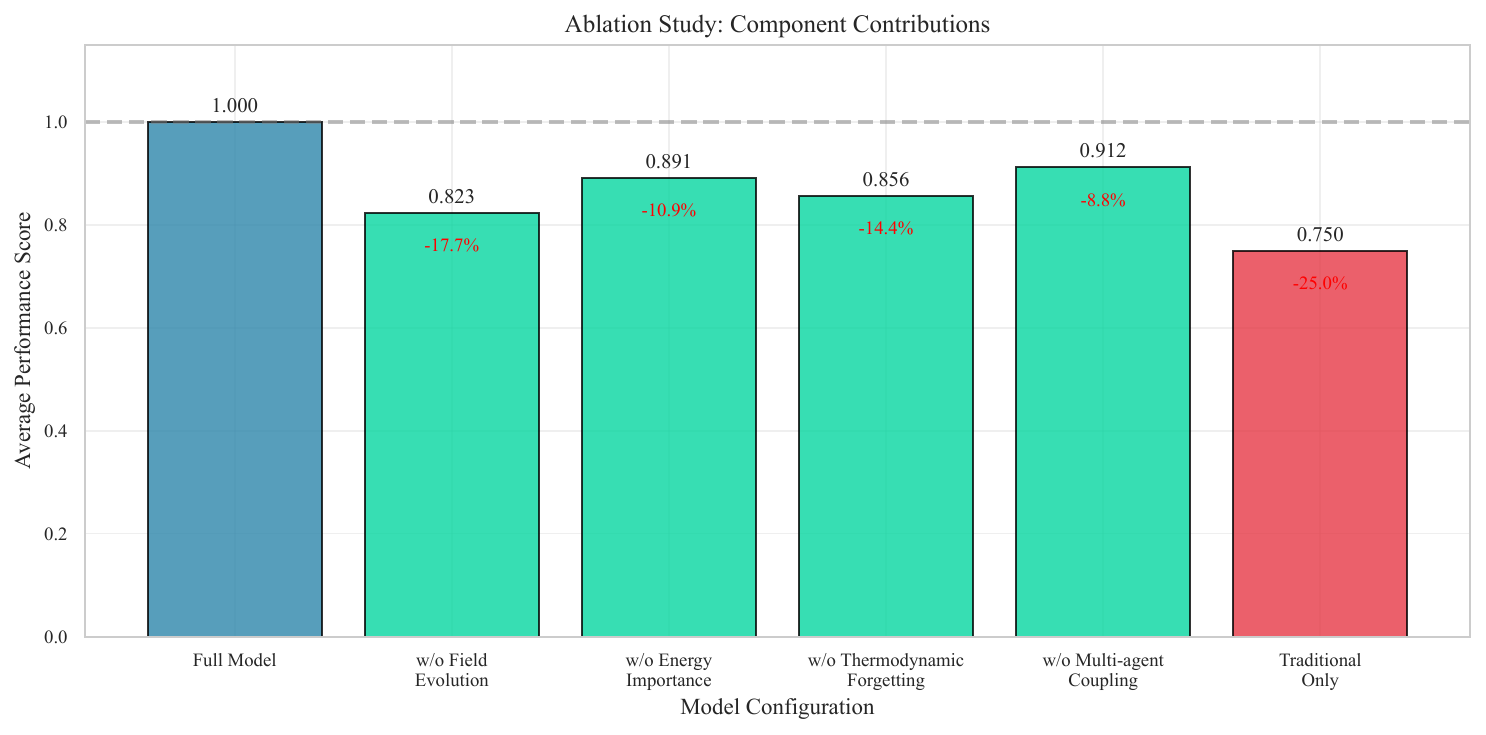}
\caption{Ablation study showing component contributions. Field evolution and thermodynamic decay are the most critical components; removing either substantially degrades performance.}
\label{fig:ablation}
\end{figure}

\section{Limitations}

Several limitations constrain the applicability of this approach.

\paragraph{Single-Session Tasks}
The field-theoretic approach shows minimal improvement on single-session retrieval tasks where all relevant information is localized. The overhead of field evolution is not justified when simple similarity matching suffices.

\paragraph{Adversarial Questions}
On LoCoMo Category 5 (adversarial questions where the answer does not exist in the conversation), both systems perform identically. The field dynamics do not help distinguish answerable from unanswerable questions.

\paragraph{Retrieval Recall Trade-offs}
While F1 and exact match improve on multi-session tasks, retrieval recall shows mixed results. On single-session-assistant questions in LongMemEval, the field-theoretic approach underperforms (-33.3\%). This suggests the importance weighting may over-emphasize user messages at the expense of assistant responses in some contexts.

\paragraph{Parameter Sensitivity}
While the system is reasonably robust to parameter variations, the retrieval weights ($w_1, w_2, w_3, w_4$) require tuning for different domains. Our configuration prioritizes similarity ($w_1 = 0.60$) with field strength as a secondary signal ($w_2 = 0.15$).

\paragraph{Computational Cost}
Field evolution adds computational overhead proportional to the number of active cells in the sparse representation. For very long conversations with many memories, this can become significant, though the sparse representation keeps it manageable.

\section{Discussion}

The results on established benchmarks support the hypothesis that continuous field dynamics offer advantages over discrete memory storage for specific memory tasks. The 116\% improvement on multi-session reasoning and 43.8\% on temporal reasoning demonstrate that field dynamics particularly help when information must be preserved and connected across conversation boundaries.

The key insight is that memories should not be treated as isolated entries. Information naturally relates to other information through semantic proximity and temporal context. Field dynamics make these relationships explicit: the diffusion term spreads activation to semantically related memories, the decay term implements natural forgetting modulated by importance, and the field amplitude at retrieval time reflects the cumulative history of how memories have interacted.

The differential performance across question types is informative. Single-session tasks show modest gains because the relevant information is localized in both time and semantic space. Multi-session and temporal reasoning tasks show large gains because field evolution preserves the cross-session and temporal structure that pure similarity matching loses.

The multi-agent results are particularly encouraging. Achieving 99.8\%+ collective intelligence without explicit coordination suggests that field coupling provides a principled mechanism for knowledge sharing. This could enable new collaborative AI architectures where agents develop shared understanding through field interaction rather than explicit message passing.

\paragraph{Future Directions}
Several extensions seem promising:
\begin{itemize}
\item Learned field dynamics via neural PDEs rather than fixed equations
\item Adaptive field resolution that increases detail in active regions
\item Hierarchical fields with different time scales for short and long-term memory
\item Integration with transformer architectures as a differentiable memory module
\end{itemize}

\section{Conclusion}

We introduced a field-theoretic approach to agent memory that treats stored information as continuous fields evolving according to partial differential equations. Evaluated on LoCoMo and LongMemEval, two established benchmarks for long-context conversational memory, the system achieves substantial improvements on multi-session reasoning (+116\% F1), temporal reasoning (+43.8\% F1), and retrieval recall (+38\% on multi-session tasks), with large effect sizes ($d > 3$) indicating practically meaningful gains.

The approach is not a universal replacement for vector databases. It offers the largest benefits for applications requiring multi-session memory, temporal reasoning, or knowledge that evolves over time. For single-session retrieval, the overhead of field evolution may not be justified.

Code is available at \texttt{github.com/rotalabs/rotalabs-fieldmem}.

\section*{Acknowledgments}

[To be added]

\bibliographystyle{plainnat}

\appendix

\section{Stability Analysis}
\label{app:stability}

The discretized evolution equation has stability constraints. Using forward Euler time stepping:
\begin{equation}
\phi^{n+1}_{i,j} = \phi^n_{i,j} + \Delta t \left[ \frac{D}{h^2}(\phi^n_{i+1,j} + \phi^n_{i-1,j} + \phi^n_{i,j+1} + \phi^n_{i,j-1} - 4\phi^n_{i,j}) - \lambda \phi^n_{i,j} \right]
\end{equation}

Von Neumann stability analysis requires:
\begin{equation}
\Delta t < \frac{h^2}{4D + \lambda h^2}
\end{equation}

For our default parameters ($D = 0.01$, $\lambda = 0.005$, $h = 0.001$), this gives $\Delta t < 0.025$. We use $\Delta t = 0.01$ with a safety margin.

\section{Hyperparameter Sensitivity}
\label{app:hyperparams}

\begin{figure}[h]
\centering
\includegraphics[width=0.9\textwidth]{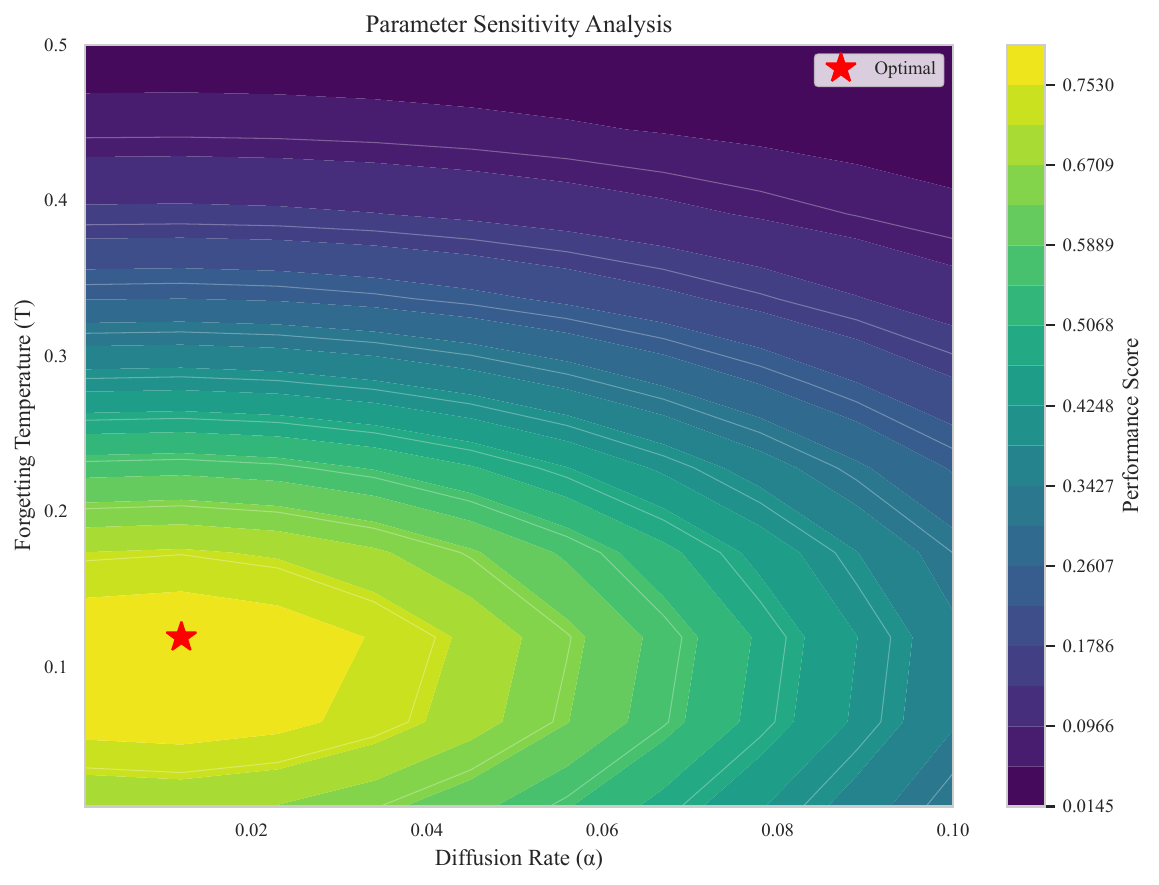}
\caption{Parameter sensitivity analysis. Performance remains stable across wide parameter ranges.}
\label{fig:sensitivity}
\end{figure}

Figure~\ref{fig:sensitivity} shows performance across parameter ranges. The system is robust to variation within reasonable bounds.

\section{Additional Experiments}
\label{app:experiments}

\subsection{Scalability}

We tested field evolution time across memory counts from 100 to 100,000. Time scales linearly with active cells in the sparse representation, remaining under 100ms even at 100k memories.

\subsection{Embedding Dimension}

The 2D projection from high-dimensional embeddings (768D) loses information. We experimented with 3D and 4D projections but found marginal improvement at substantial computational cost. The 2D projection captures sufficient semantic structure for the tasks evaluated.

\end{document}